\documentclass{article}

\usepackage[sectionbib,square]{natbib}

\usepackage{natbib}

\PassOptionsToPackage{hyphens}{url}\usepackage{hyperref}

\usepackage[preprint]{modified_neurips_2020}

\usepackage{bm}
\usepackage{graphicx}
\usepackage{subcaption}
\usepackage{url}            
\usepackage{enumitem}
\usepackage{booktabs}       
\usepackage{microtype}      
\usepackage{algorithm,algorithmic}

\usepackage{verbatim}

\usepackage{amsthm,amsmath,amssymb,amsfonts,bbm,nicefrac}

\usepackage{import}
\usepackage{float}
\usepackage{tikz, pgfplots,tabularx}
\usepackage{calc}
\usepackage[normalem]{ulem}

\usepgfplotslibrary{fillbetween}
\usetikzlibrary{positioning, matrix}

\theoremstyle{definition}
\newtheorem{theorem}{Theorem}

\newtheorem{lemma}{Lemma}
\newtheorem{proposition}{Proposition}

\newtheorem{definition}{Definition}

\theoremstyle{remark}

\newenvironment{proofsketch}{%
  \proof}{\endproof}

\newif\ifedits
\editsfalse

\DeclareRobustCommand{\bb}[1]{\mathbb{#1}}

\DeclareMathOperator{\Var}{Var}

\definecolor{roc1}{HTML}{8D0026}
\definecolor{roc2}{HTML}{83BEDE}
\definecolor{fill}{HTML}{D5D5D5}
\definecolor{feas}{HTML}{BD0000}

\usepackage{setspace}

\begin{document}
\floatname{algorithm}{Algorithm Part}

\title{\onehalfspacing \Large  Affirmative Action vs. Affirmative Information}

\author{%
  Claire Lazar Reich   \\  Stanford University  \\ 
}

 \doublespacing
 
\maketitle

\begin{abstract}
\onehalfspacing
Critical decisions in hiring, college admissions, and credit lending  are guided by predictions made in the presence of uncertainty. While uncertainty imparts   errors across all demographic groups, this paper shows that the types of errors vary systematically: Groups with higher average outcomes are typically assigned higher false positive rates, while those with lower average outcomes are assigned higher false negative rates. 
We characterize the conditions that give rise to this disparate impact and explain why the intuitive remedy to omit demographic variables from datasets does not correct it.  
Instead of data omission, this paper examines how data acquisition can broaden access to opportunity.  The strategy, which we call  “Affirmative Information,” could  stand as an alternative to Affirmative Action.

 \end{abstract} 


\newpage

\section{Introduction} 
\label{intro}

Decision-makers that match individuals to opportunities wrestle with uncertainty and inevitably make  mistakes. 
Empirical studies of foster care assignments,  criminal bail decisions, and hospital treatments have uncovered real-world cases in which 
 prediction errors   disproportionately affect the 
 most disadvantaged subpopulations being screened. 
   \citep{angwin2016machine, arnold2022measuring, 
   eubanks2018child, obermeyer2019dissecting}. 
These patterns cause systematic differences in access   to opportunities that are needed and deserved.
   
To  explain  biased outcomes,  intuition often guides us to search for  bias in the data that drive the predictions. 
This paper, however, shows that bias enters predictions not only through included data, but also through the absence of data.    
We explain how uncertainty itself complicates   efforts to identify \textit{equally qualified} candidates from different demographic groups and thereby generates disparate impacts.

In particular, we consider screening decisions in which applicants are either granted or denied an opportunity, such as a loan, based on predictions of their eventual outcomes, such as loan repayment. 
We show that while uncertainty affects the accuracy of predictions for all subpopulations in the data, the \textit{types} of errors often vary systematically. 
Applicants from lower-average groups tend to be assigned higher false negative rates and those from higher-average groups tend to be assigned higher false positive rates. 
As a result, applicants from lower-average groups are less likely to be accepted conditional on their true ability. 
We call this effect ``the disparate impact of uncertainty" and characterize why it often arises in practice, despite decision-maker's efforts to make   predictions fairly. 

The reason that the disparate impact of uncertainty  arises so often is that it can  emerge even from facially neutral prediction methods.
We show that the widespread practice of stripping demographic data from datasets is  not a remedy.
This distinguishes the disparate impact of uncertainty from traditional statistical discrimination, in which differences in access to opportunity can be traced back to the  decision-makers' choice to weigh demographic characteristics \citep{aigner1977statistical}. 

Rather than relying on demographic characteristics, the disparate impact of uncertainty emerges due the mechanistic force of regression toward the mean. In the presence of uncertainty, regression toward the mean causes the overall distribution of predicted ability to contract toward the  mean ability, but we show that predictions for different groups contract toward different means when the underlying data  is correlated with group membership. 
Uncertainty pulls predictions of applicants from lower-mean groups toward lower values, causing the distribution of predictions given true ability to  vary systematically by group. 

Since the variables included in prediction models often explain \textit{how}   demographic groups are disadvantaged, the disparate impact we characterize here regularly arises in practice. In the lending context, for example, we rarely suppose that demographic characteristics actually drive observed differences in group repayment rates. Instead, demographic characteristics are correlated with  financial characteristics that do drive repayment ability. When lenders  include those financial characteristics in their models, they  inadvertently create opportunities for uncertainty to generate  disparate impacts across demographic groups.

Fortunately, it is possible for decision-makers to eliminate the disparate impact of uncertainty. To do so, we prove that at least one of the following must be true: (1) members of the lower-mean groups must be over-represented among positive classifications, OR (2) their predictions must be more accurate than those of higher-mean groups. 
The second prong of this mathematical result motivates us to propose a practical method to broaden access to opportunity across  qualified applicants: 
acquire additional data about individuals in disadvantaged groups. 
We call this strategy ``Affirmative Information.''
It could  promote both fairness and accuracy in the wake of the United States Supreme Court's historic decision to strike down Affirmative Action in \textit{SFFA v. Harvard} and \textit{SFFA v. UNC}.

In contrast to our proposal of data \textit{enrichment}, many prediction models in use today seek to achieve fairness through data \textit{omission}. 
This paper therefore supports the growing evidence in both economics and computer science showing that blinding algorithms to demographic characteristics is an ineffective approach to fair prediction. Within this literature, \citet{10.1145/3328526.3329621, dwork2018decoupled, rambachan2020economic}
advance theory showing that fairness considerations are best imposed after estimating the most accurate predictions, rather than before, and \citet{kleinberg2018algorithmic, reich2020possibility} demonstrate this principle empirically in real-world settings. 
Our paper aligns with recent evidence  that data collection can serve as an integral part of fair prediction, as advanced by \citet{chen2018my}, and that fairness interventions can actually enhance accuracy and overall utility, as shown in \citet{kleinberg2018selection, Emelianov_2020}.


While we focus on  equal opportunity in this paper, it is important to acknowledge that there are multiple notions of fair prediction.
 Among them is accuracy on the individual level \citep{dwork2012fairness}. To safeguard  this criterion, we do not propose eliminating error disparities by intentionally reducing the accuracy of predictions for individuals in higher-mean groups. Rather, 
we consider the strategy of using the immediately available data, as  in \citet{kleinberg2018algorithmic}, supplemented with further data acquisition efforts to better identify  qualified individuals from  lower-mean groups. 

To summarize the paper's primary findings: 
\begin{enumerate}
\item  Predictive uncertainty imparts a disparate impact: even when demographic variables are omitted from datasets, applicants from lower-mean groups are often more likely to be rejected than those from higher-mean groups \textit{conditional on their true ability}.
\item To eliminate the disparate impact, decision-makers must over-select members of lower-mean groups OR     predict their outcomes more accurately. 
\item Affirmative Information is a promising avenue to  broaden  access  to opportunity.  
\end{enumerate}

The paper proceeds as follows. 
Section 2 grounds the discussion with a brief example of a facially neutral prediction scheme that generates disparate impacts across equally-qualified loan applicants. 
 Section 3 presents the theory that explains why and when the disparate impact of uncertainty emerges in practice.
 It also introduces a model  to relate the results here to the canonical labor discrimination paper by \citet{aigner1977statistical}.
Section 4 empirically compares two methods  to overcome the disparate impact of uncertainty: Affirmative Action and Affirmative Information. We show that in settings with selective admissions standards, Affirmative Information can more effectively increase the probability that qualified individuals in disadvantaged groups are granted opportunity. Depending on the costs of data acquisition, it can simultaneously yield higher return to the decision-maker and create natural incentives to admit more applicants.

\section{Motivating example} \label{motiv}

Motivating this study is the observation that predictions can systematically favor individuals from high-mean groups even when they do not discriminate by group. 
In this section, we empirically illustrate this  phenomenon in a credit-lending setting. 

We consider a lender that uses risk scores to screen  loan applicants. We suppose the designer of these risk scores is tasked with providing high-quality predictions that are fair to demographic subgroups, one highly educated $H$ (more than a high school degree) and another less educated $L$ (at most a high school degree). The lender avoids discrimination by purging the dataset of sensitive characteristics, including demographic variables, leaving only  financial variables deemed relevant for loan repayment.

\begin{figure}[!b] \vspace{.5cm}
  \centering
  \includegraphics[width=\textwidth]{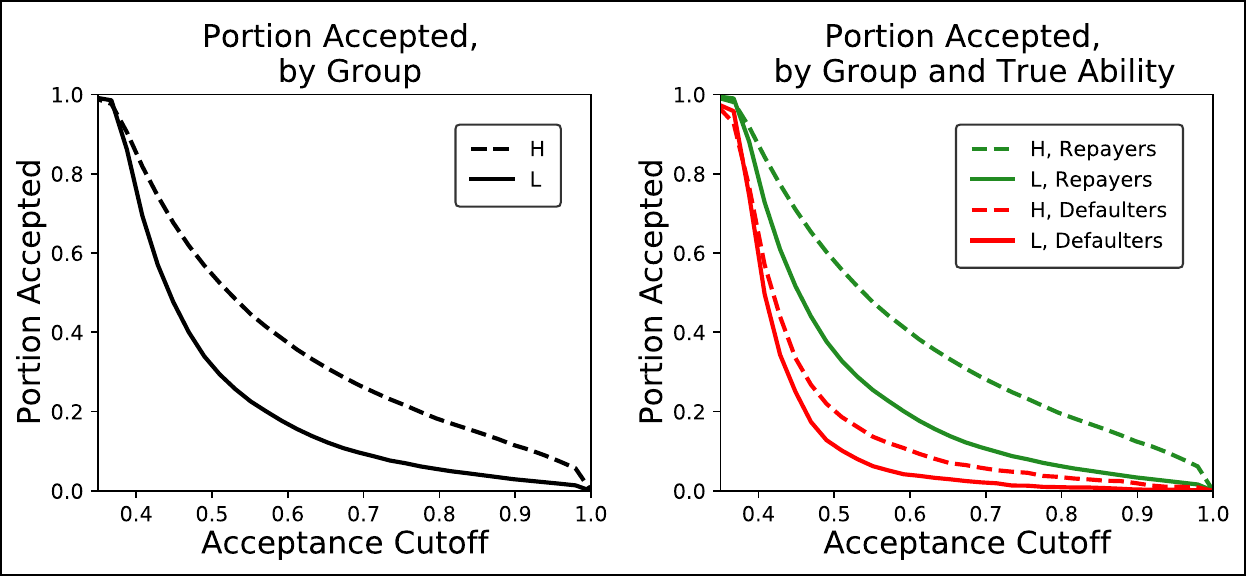} 
  \caption{\label{motivation-fig} \small (a) Members of the higher-mean group are more likely to be classified as creditworthy. (b) Members of the higher-mean group are more likely to be classified as creditworthy, even when we condition on true ability. While the predictions are group-blind and there was no disparate treatment, there is disparate impact.\vspace{.5cm}}
\end{figure}

We demonstrate this task on real financial data from the Survey of Income and Program Participation (SIPP), defining the outcome of interest as successfully paying rent, mortgage, and utilities in every month of 2014 \citep{sipp_cite}. We estimate the probability of successful payment for each adult in the dataset using logistic regression on three predictors: total assets, total debt, and monthly income.  The lender could then translate these predicted probabilities into  loan approvals and rejections at any cutoff of their choosing.  

Figure \ref{motivation-fig} depicts loan access by group, given all possible cutoffs on the horizontal axis. In the left panel, we see that less-educated individuals are less likely to be classified as creditworthy at all cutoffs. This fact alone is not alarming, since it could reflect that the predictions have  identified legitimate differences in repayment abilities across educational groups. However, in the right panel, we show that less-educated individuals are also less likely to be classified as creditworthy \textit{even when we condition on their eventual repayment outcome}. That is, they have lower true and false positive rates than high-educated applicants. In practice, this means that a decision-maker guided by these predictions would be more likely to grant a loan to a creditworthy $H$ applicant than a creditworthy $L$ applicant, and also more likely to grant a loan to a non-creditworthy $H$ applicant than a non-creditworthy $L$ applicant.

As we will prove in Section 3, the disparate impact  identified here  arises when group affiliation is correlated with variables included in the prediction model. 
It regularly emerges in practice because we can often trace the disadvantage of
disadvantaged groups through a multitude of variables, including those that are essential to
predicting the outcomes we care about.

\section{Results} \label{results-sec}


\subsection{Statistical results without distributional assumptions} \label{when-sec}

 This section presents  general statistical results applicable to
 settings in which   applicants are screened  for opportunities. 
 We begin with the usual observation that uncertainty affects predictions of applicant ability via regression toward the mean. 
We show how the effect varies across groups of applicants when i) average abilities differ by group and 
ii) the  data guiding  predictions is correlated with group membership.
Equally-qualified individuals from different groups can then be expected to have  different predicted abilities.  
We  identify necessary conditions to ensure that equally-qualified individuals from different groups are accepted at equal rates: either applicants from the lower-average groups must be screened with greater accuracy or they must be over-represented among acceptances.  
At least one of these conditions must be met to achieve equal opportunity in practice. 

Consider applicants  each with   ability $A$.
The applicants belong to groups $G\ \in \{ L, H\}$ that differ in average ability, so that $\mu_L < \mu_H$ where $\mu_G \equiv \mathbb{{E}}[A|G]$.
A decision-maker seeks to estimate $A$ using available applicant data $X$. 
At no point does the decision-maker observe protected group membership $G$. However, the data $X$ may be correlated with group membership.
The decision-maker estimates the conditional expectation function (CEF) $\bb E [A|X]$ in the model
\begin{equation} \label{pop-model}
A = \bb{E} [A | X ] + \varepsilon,
\end{equation}
where $\varepsilon$ is a residual that is uncorrelated with any function of $X$\footnote{by the CEF-Decomposition Property \citep{angrist_mostly_2008}}, and  $ \bb E[A|X]$ yields the best predictions given the available data\footnote{i.e., the CEF minimizes mean squared error, by the CEF-Prediction Property \citep{angrist_mostly_2008}}.

\begin{figure}
   \makebox[\textwidth][c]{\includegraphics[width=1\textwidth]{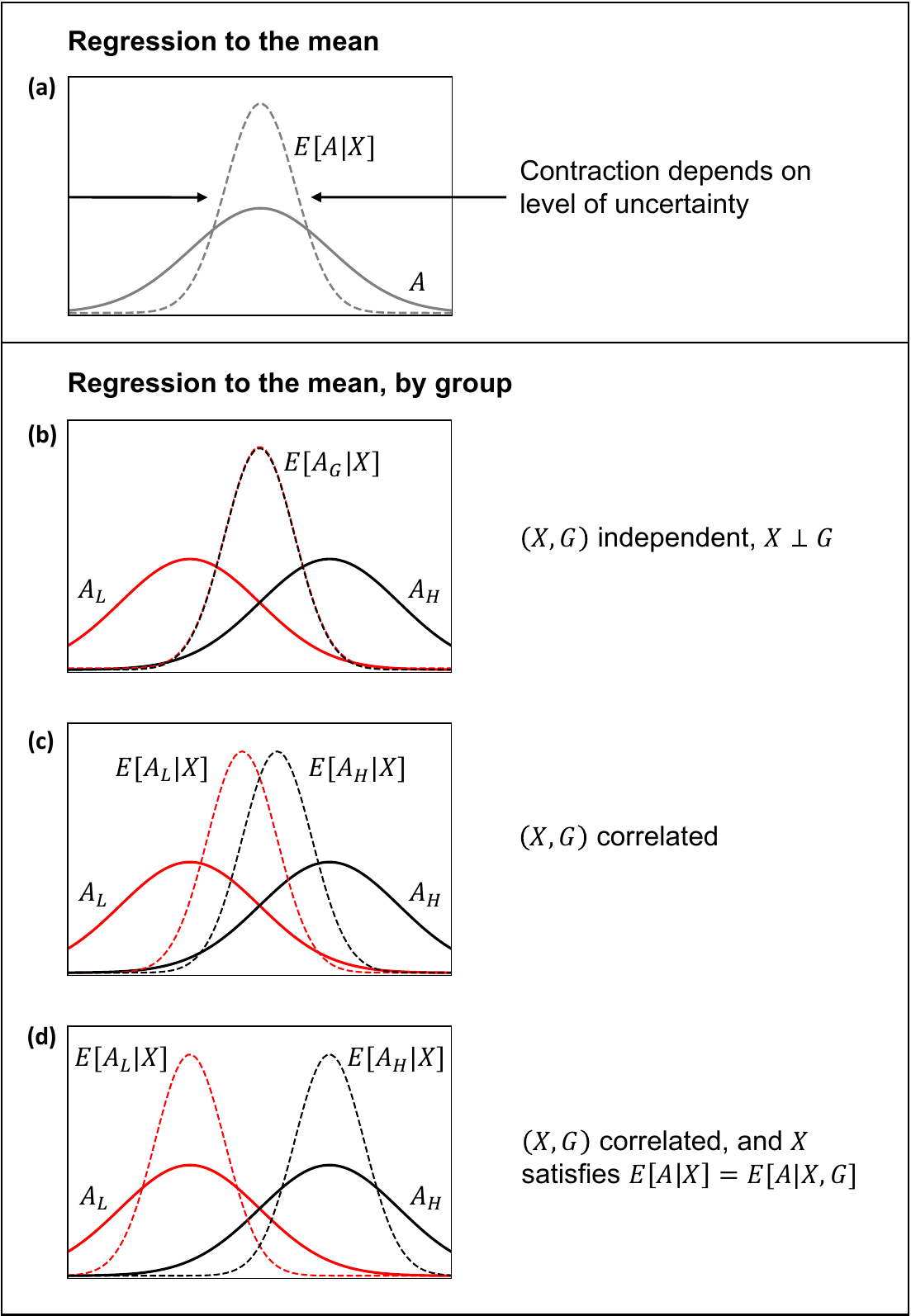}}  
  \caption{\label{cef-contraction-fig} \small (a) Regression toward the mean centers predictions $\bb E[A|X]$ at the mean of ability $A$, with smaller variance. Separately considering predictions by group, however, reveals that the effect of regression toward the mean depends on the correlation between $(X,G)$. (b) If $X \perp G$, the predictions have the same distribution. (c) If $(X,G)$ are correlated, then the mean predictions will differ by group. (d) Furthermore, if $X$ actually explains the relationship between $(A,G)$, so that $\bb E [A|X] = \bb E [A|X,G]$, then predictions are centered at group-specific means. Note that the figure depicts normal distributions for illustration, but no normality assumption is required for any statistical result in Section \ref{when-sec}. }
\end{figure}

First, note that the predictions given by the CEF follow regression toward the mean, since they are centered at the mean ability\footnote{by the law of iterated expectations, $\bb E [\bb E [A|X]] = \bb E [A] $} $\bb E[A]$, with variance smaller\footnote{by  Eq. (\ref{pop-model}), $\Var [\bb{E} [A | X] ] \leq  \Var [\bb{E} [A | X ] + \varepsilon ] = \Var[A] $} than $\Var [A]$. An illustration of contraction toward mean ability is depicted in Figure \ref{cef-contraction-fig}(a). As a result of the contraction,   high-ability individuals are expected to have above-average predictions, but not as high as their true ability. Similarly, low-ability individuals are expected to have below-average predictions, but not as low as their true ability. 
The magnitude of the contraction toward the mean  depends on  how well $X$ describes variation in $A$: the CEF is more concentrated at the mean in  high-uncertainty settings  where $X$ weakly relates to $A$. 

We can now explain how uncertainty impacts applicants  differently by group:
The contractions experienced by the groups  vary systematically depending on the relationship between $G$ and the underlying data $X$.
We depict three examples in Figure \ref{cef-contraction-fig}. 
If $X$ and $G$ are independent, the predictions follow the same distribution across groups as seen in panel (b). 
If $X$ and $G$ are not independent, however, then the predictions will be distributed as contractions toward \textit{group-dependent} means.  
This is formally shown in the next proposition.
When $X$ partly explains groups' variation in ability, so that group $L$ has a relatively greater density of applicants with $X$s associated with lower ability, then the $L$ predictions will have a lower mean.
Furthermore, if $X$ fully explains the groups' variation in ability, then  the best predictions will be centered exactly at the group-specific averages of ability. The partial and full cases are illustrated in panels (c) and (d) respectively.

\begin{proposition} \label{cef-is-contraction}
Consider the distribution of the best predictions $\bb E[A|X]$ for group $G=g$. Its variance is smaller than the variance of ability for $g$ and its mean depends on the joint distribution of $(A,X,G)$. 
If $X$ fully explains the relationship between $A$ and $G$,
so that $\bb E [A|X ] = \bb E [A|X,G]$, then the mean is exactly equal to $\mu_g$.
\end{proposition}

\begin{proof}
The variance of the best predictions for $g$ is bounded by the variance of ability, since $\Var [\bb{E} [A | X] | G=g ] \leq  \Var [\bb{E} [A | X ] + \varepsilon | G=g] = \Var[A |G=g] $, where the equality follows from (\ref{pop-model}). 

 The mean of the best predictions for $g$ is an average over $\bb E[A|X]$ weighted by the density of $X$ given $g$.  In particular, if we denote $n$ as the dimensionality of $X$ and $f_{X|G}(\bm{x}|g)$ as the conditional density of $X$ given $G=g$, then
$\bb E [  \bb E [A|X]  | G=g] = 
\int_{\bb R^n} \bb E [A|X=\bm{x} ] f_{X|G}(\bm{x}|g) d^n \bm{x}.$ 
The mean is  smaller when $g$ has a greater density of $X$s associated with low $A$. 

We can  compute the mean in closed form if  $\bb E [A|X ] = \bb E [A|X,G]$. By first applying mean independence and then applying the tower property of conditional expectation, we have $\bb E[ \bb E [A|X] | G=g]  = \bb E[ \bb E [A|X,G] | G=g] = \bb E [A|G=g] \equiv \mu_g$, so the distribution of the best predictions for $g$ is centered at the mean ability of applicants in $g$.
\end{proof}

To summarize, uncertainty impacts predictions differently by group. First, predictions of applicants in lower-mean groups will have lower means so long as $X$ contains variables that explain at least some of the variation in $A$ across groups. Then, uncertainty contracts predictions for each group closer to those group-specific means. 
We can generally expect that the presence of uncertainty causes  two applicants --- with equal true ability but from different groups --- to vary in their predicted ability.  

Decision-makers who use  predictions to identify which applicants to accept and reject   can then inadvertently generate disparate impacts.  
To see this, 
suppose a decision-maker wishes to identify applicants that are qualified for an opportunity. 
Let $Y \in \{ 0,1\}$ denote the acceptance decision they would make about an applicant if they had full certainty, i.e., $Y$ denotes whether the applicant is truly qualified.\footnote{For instance, $Y=1$ could correspond to applicants that would make timely mortgage payments over a loan period, would not recidivate if granted bail, or would graduate an academic program in good standing.} 
Suppose qualification rates are lower in group $L$ than in $H$, with   $0< \mu_L < \mu_H<1 $ where $ \mu_G = \bb P [Y=1|G ]$.
Meanwhile, let  $\hat y \in \{ 0,1\} $ be the decision-maker's   choice to reject or accept the applicant given the presence of uncertainty.
We make \textit{no assumptions} about whether  $\hat y$ is based on data that is correlated with group membership.

The ``disparate impact of uncertainty'' arises if  qualified members of $L$ are rejected at higher rates than  qualified members of $H$.\footnote{With fairness in mind, we focus on false negative rates (FNRs) because they represent barriers to   opportunity for qualified applicants.  False positive rates (FPRs), by contrast,  represent probabilities that unqualified applicants access the opportunity, and equating these by group may be unmotivated from a fairness perspective \citep{hardt2016equality}. For mathematical completeness, we include results  on balancing both FNRs and FPRs in the  appendix.}

\begin{definition}
The disparate impact of uncertainty is present when $\text{FNR}_L >  \text{FNR}_H $, where  $\text{FNR}_G$ is the group-specific false negative rate $ \bb P [\hat y = 0 | Y=1, G] $.
This  is mathematically equivalent to $\text{TPR}_L < \text{TPR}_H$, where $\text{TPR}_G$ is the group-specific true positive rate $ \bb P [\hat y = 1 | Y=1, G] $. 
\end{definition}

The following lemma guarantees that $\hat y$ will  exhibit the disparate impact of uncertainty if
i)  the share of accepted individuals from $L$ is no greater than the share of truly qualified individuals from $L$, and 
ii) classifications of $L$   are no more accurate than  those of $H$. 
A proof sketch is included, and interested readers are referred to a detailed proof in the appendix.\footnote{We assume there is predictive uncertainty: In particular,  both groups' true and false positive rates are bounded by 0 and 1, so that for all $G$,  $\bb P [\hat y = 1 | Y=1, G] ,  P [\hat y = 1 | Y=0, G] \in (0,1)$. \label{footnote-rate-assumption}}
  
\newpage

\begin{lemma} \label{accuracy-thm}
$\hat y $ exhibits the disparate impact of uncertainty if
\begin{itemize}
\item[(i)]  it does not over-represent members of $L$  among positive classifications: 
\begin{equation*}
\mathbb{P}[G=L | \hat y =1] \leq \mathbb{P}[G=L | Y =1],
\end{equation*}
\item[(ii)] and its  positive or negative predictions for members of $L$ are less accurate  than those of $H$: 
\begin{equation*}
\text{PPV}_L < \text{PPV}_H  \text{ or }   \text{NPV}_L \leq \text{NPV}_H,
\end{equation*}
where $\text{PPV}_G \equiv \mathbb{P}[Y=1| \hat y=1, G]$ and 
 $\text{NPV}_G  \equiv \mathbb{P}[Y=0| \hat y=0, G] $ are the group-specific positive and negative predictive values.
\end{itemize}
\end{lemma}
\noindent

\begin{proofsketch}
Let  $\hat \mu_G$ denote the group-specific positive selection rate, $\bb P [\hat y =1 |G]$.  If (i) holds, then 
\begin{equation} \label{overrep-cond-rewritten}
 \frac{\mu_H  }{\mu_L} \leq \frac{\hat \mu_H }{\hat \mu_L}.
\end{equation}
Suppose (ii) also holds, separately considering the cases $\text{PPV}_L < \text{PPV}_H$ and $\text{NPV}_L \leq \text{NPV}_H$.

\noindent
Case 1: Rewriting (\ref{overrep-cond-rewritten}) gives $\frac{\hat \mu_L}{\mu_L}  \leq   \frac{\hat \mu_H}{\mu_H}$, and combining with $\text{PPV}_L < \text{PPV}_H$, we get 
\begin{equation*} \label{tpr-bayes}
 \text{PPV}_L \frac{\hat \mu_L}{\mu_L} <  \text{PPV}_H \frac{\hat \mu_H}{\mu_H}. 
\end{equation*} 
This is  $\text{TPR}_L < \text{TPR}_H$ rewritten with Bayes' rule. 

\noindent
Case 2: Note that  (\ref{overrep-cond-rewritten}) implies $\hat \mu_L < \hat \mu_H$, since $\mu_L < \mu_H$. Combining with $\text{NPV}_L \leq \text{NPV}_H$,  
\begin{equation*}
(1-\text{NPV}_L) \frac{1-\hat \mu_L}{\mu_L} > (1-\text{NPV}_H) \frac{1-\hat \mu_H}{\mu_H}. \label{fnr-bayes} 
\end{equation*}
This is $\text{FNR}_L > \text{FNR}_H$ rewritten with Bayes' rule, where $\text{FNR}_G$ is the group-specific false negative rate $\bb P [\hat y = 0 | Y=1, G]$. Because false negative  and true positive rates are related by $\text{FNR} = 1 - \text{TPR}$, we have $1- \text{TPR}_L > 1- \text{TPR}_H \implies \text{TPR}_L < \text{TPR}_H$.  
 \end{proofsketch}

The following theorem presents the contrapositive. It identifies necessary conditions to eliminate the disparate impact in all settings in which uncertainty is present and qualification rates differ by group:

\begin{theorem} \label{nec-conditions-result}
If $\hat y$ does not exhibit the disparate impact of uncertainty, then it must  over-represent members of $L$ among positive classifications or predict their outcomes more accurately:
\begin{itemize}
\item[(i)]  $\mathbb{P}[G=L | \hat y =1] > \mathbb{P}[G=L | Y =1]$, or
 \item[(ii)]  $\text{PPV}_L \geq \text{PPV}_H  \text{ and }  \text{NPV}_L > \text{NPV}_H$.
 \end{itemize}
\end{theorem}

 Lemma \ref{accuracy-thm} and Theorem \ref{nec-conditions-result} hold in general, as 
their proofs do not depend on assumptions about how the decision-maker arrives at $\hat y$. 
Therefore, Theorem \ref{nec-conditions-result} provides practical guidance across a wide range of settings. It establishes that there are only two ways to overcome the disparate impact of uncertainty: The decision-maker can   i)  design prediction schemes that over-select members of a group with lower qualification rates, such as by setting more lenient admission criteria  for members of that group.\footnote{Note that the over-representation condition may also be achieved in settings where admission criteria are lenient for all groups. For example, suppose  $\mu_L = .1$, $\mu_H=.9$, half the population is $L$, and virtually all applicants are admitted. Then 50\% of positive classifications are from $L$, though only 10\% of qualified individuals are from $L$.} 
Alternatively, the decision-maker can  ii) increase the accuracy of predictions for this group, such as by investing in additional data acquisition.



\subsection{A stylized model with distributional assumptions} \label{model-theory-sec}

In this section,  we introduce additional assumptions in the form of a stylized model to show how the disparate impact of uncertainty emerges from a generalization of statistical discrimination.
These results  expand the canonical insights of  \citet{aigner1977statistical}  to settings where decision-makers screen applicants based on an arbitrary set of variables, rather than
 on an unbiased test score and a demographic variable.
In this generalization,  the disparate impact discovered by \citet{aigner1977statistical} can emerge even when decision-makers are barred from using demographic variables. We show the disparate impact can be eliminated by increasing the  explanatory power of the predictors for the affected group.


Suppose an accuracy-maximizing lender  uses  data $X$ to predict repayment ability $A$. For this model, unlike the previous section, we add another assumption: 
Assume that $X$ explains variation in $A$ across demographic groups $G$ so that $\bb E [A|X ] = \bb E [A|X,G]$. This holds trivially if  $G \in X$. But it could also hold if $X$ contains financial data that explain the lower repayment rates of the disadvantaged group.

As in Eq.  (\ref{pop-model}) from the previous section, applicant ability $A$ can be decomposed into a signal that is explained by the data $X$ and a residual that is unexplained by $X$. 
We can write this decomposition for group $G=g$ as
\begin{equation} \label{ability-decomposition}
A_g = S_g + \varepsilon_g, \\
\end{equation}
where the signal $S_g \equiv \bb E [A_g | X_g]$ is the most accurate prediction of ability given the data. 

Suppose that the lender observes  $S_g$ and accepts  all applicants that have $S_g$ above a cutoff of her choosing.
The choice of cutoff depends on her relative valuation of false positive and false negative classifications. 
In a setting where false negatives are more costly, the cutoff will be lower, and in a setting where false positives are more costly, the cutoff will be higher. 
We call low-cutoff settings ``lemon-dropping markets" and high-cutoff settings ``cherry-picking markets," as in \citet{bartovs2016attention}.

For model tractability, we assume as in \citet{aigner1977statistical} that
$S_g$ and $\varepsilon_g$ are bivariate normal. 
 Our model is therefore specified by   signals $S_g \sim \mathcal{N}(\mu_g, \sigma^2_{S_g})$,  residuals\footnote{By the assumption that $\bb E [A|X ] = \bb E [A|X,G]$ and   Proposition \ref{cef-is-contraction}, both groups' residuals must be mean-zero.} $\varepsilon_g \sim \mathcal{N}(0, \sigma^2_{\varepsilon_g})$, and ability  $A_G \sim \mathcal{N}(\mu_g, \sigma^2_{S_g} + \sigma^2_{\varepsilon_g} )$. 
  We can use the bivariate normal distribution assumption to calculate the expected signal for a member of group $g$ with ability $a$. Derived in the appendix, it is
\begin{equation}
\bb E[S_g | A_g = a]  = (1-\gamma_g) \mu_g + \gamma_g a \text{, where } \gamma_g \equiv    \frac{\sigma_{S_g}^2}{\sigma_{S_g}^2+ \sigma_{\varepsilon_g}^2}. 
\end{equation}
Therefore, the expected signal is a weighted average of the true ability and the group mean. 
The weight $\gamma_g$ on the true ability reflects the portion of $A_g$'s total variance that is captured by the observable signals $S_g$, and the weight $1-\gamma_g$ on the mean reflects the unexplained variance due to the presence of uncertainty. As $\gamma_g$ increases, the signal is expected to closely track true ability. 
As $\gamma_g$ decreases, however, the signals are expected to bunch more tightly about the group mean.  

 An immediate consequence is that an individual from a lower-mean group $L$ with true ability $a$ can generally be expected to have a different predicted ability than an individual from a higher-mean group $H$ with the same ability $a$.
 
We next consider two cases, one demonstrating the disparate impact of uncertainty and the second correcting  it. The first case equates the observed variance in ability ($\gamma_L = \gamma_H$), while the second explores the effect of increasing the observed variance for the $L$ group ($\gamma_L > \gamma_H$). In each case, the best  estimate of ability for each group is given by the signal.
 However, in the latter case, the estimate is made more precise for $L$.

\begin{figure}
   \makebox[\textwidth][c]{\includegraphics[width=\textwidth]{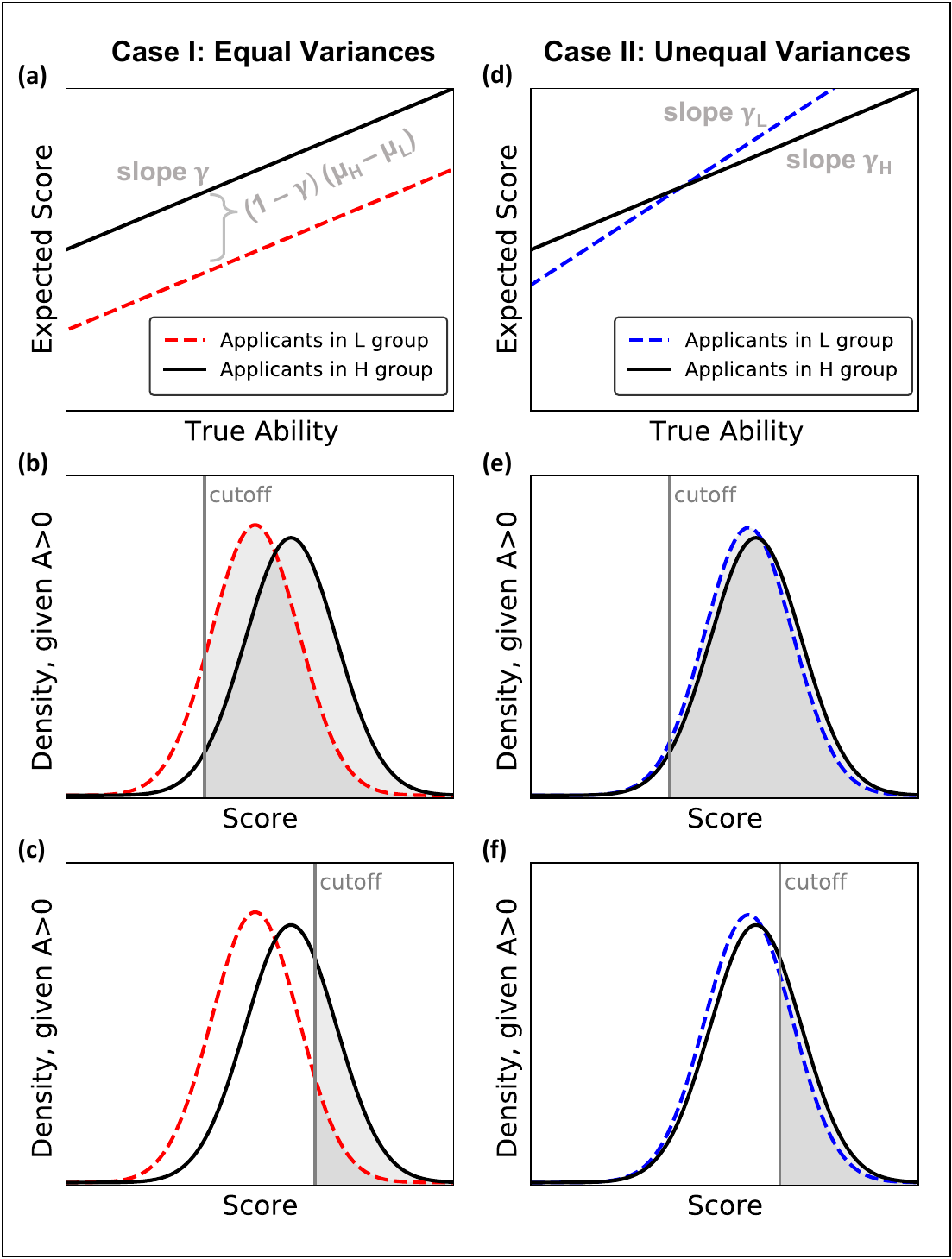}}  
  \caption{\label{panel-fig} \small Plots in left column illustrate the case with equal variance signals, and plots in right column are based on higher-variance signals for the $L$ group. In (a), we see that expected signals conditional on true underlying ability are systematically higher for $H$ members. In  (b) and (c), we plot the distribution of signals for qualified applicants ($A>0$) in a lemon-dropping and cherry-picking market, respectively. The group-specific true positive rates are given by the portion under each curve that is shaded. Members of $H$ have higher true positive rates, and the difference is particularly stark in the cherry-picking example. The imbalance is alleviated when the  $L$ signals become more informative, as seen in (d)-(f).  }
\end{figure}

 \subsubsection{Equally informative signal}

First consider a setting with 
 $\gamma^L = \gamma^H \equiv \gamma$. We depict expected signals conditional on true underlying ability in Figure \ref{panel-fig}(a). For each level of ability, individuals in the $H$ group are expected to have higher observable signals. The gap between their expected signals and those of the $L$ group is seen to increase  in the differences between their means $(\mu_H - \mu_L)$ and the extent of the uncertainty, $(1-\gamma)$. 

Since the lender applies the same cutoff rule to all applicants, the systematic disparities in signals  propagate into systematic disparities in  the ultimate acceptance decisions. 
We formally show this by defining applicants' true qualification $Y = \mathbbm 1\{A>0\}$ and deriving  the conditional distribution of $S$ given  $Y$.\footnote{The conditional distribution of $S|\mathbbm 1\{A>0\}$ is normal. Derivation of its mean and variance is provided in the appendix.}
In Figure \ref{panel-fig}(b) and \ref{panel-fig}(c) we illustrate the distributions of signals of qualified individuals,  given two examples of lender cutoffs\footnote{The distributions depicted are based on parameter values $(\mu_H, \sigma_{S_H}^2, \sigma_{U_H}^2) = (0, 1, 1)$ and $(\mu_L, \sigma_{S_L}^2, \sigma_{U_L}^2) = (-1, 1, 1)$. The lemon-dropping and cherry-picking cutoffs depicted are $-1$ and $+1$, respectively.}. 
Because the true positive rate is the portion of qualified individuals ultimately granted a loan, it is straightforward to visualize it on the plot: each group's true positive rate is given by the portion of the area under its curve that is shaded, corresponding to those applicants with a signal above the cutoff. We  observe that the true positive rates are higher for the $H$ group in both lemon-dropping and cherry-picking markets, and that the effect is most stark when the lender  cherry-picks applicants at the top of the distribution, as in Figure \ref{panel-fig}(c).

 \subsubsection{More informative $L$ signal}
 
Next consider the case where the signal captures a greater portion of the variance in ability for the $L$ group than for the $H$ group, that is when $\gamma^L > \gamma^H$. Then imbalances in the true positive rate are reduced. The expected signals conditional on true underlying ability are plotted in Figure \ref{panel-fig}(d), where we see that at low levels of underlying ability, $L$ members have lower expected signals than $H$ members, whereas they have higher expected signals at higher levels of ability. 

Meanwhile, the distributions of the signals of qualified individuals are plotted in Figures  \ref{panel-fig}(e) and \ref{panel-fig}(f)\footnote{The plots in Figures \ref{panel-fig}(e) and \ref{panel-fig}(f) are constructed by changing the $L$ variances to $(\sigma_{S_L}^2, \sigma_{U_L}^2) = (1.6, 0.4)$.}. 
 After increasing the signal variance for $L$, we see that  the shaded portion under each groups' curve is approximately the same, signifying that the groups' true positive rates are approximately the same.

\newpage
 
 \subsubsection{Connection to  other discrimination models}

\noindent \textbf{Biased measurement models}

Group-blind prediction is known to underrepresent disadvantaged groups when decision-makers estimate applicant ability  with implicit bias in means or with different levels of variance. 
 \citet{kleinberg2018selection, Emelianov_2020} characterize these phenomena in  cases where groups have the same underlying  ability distributions. 

By contrast, here we consider underrepresentation emerging from unbiased measurement of two groups' \textit{differing} ability distributions. 
We are motivated by practical cases where past disadvantage makes it harder for some applicants to succeed if they were screened in by the decision-maker.    
For example, members of  a disadvantaged group may have relatively lower ability to perform well in college after coming from underfunded high schools, or lower ability to pay back loans due to lower access to high-paying jobs.   
In  the corresponding decisions to admit students to college or grant applicants loans, we have shown that equal levels of predictive uncertainty across groups can lead not only to general underrepresentation of disadvantaged groups, but actually to underrepresentation of their  \textit{equally-capable applicants}. 

\noindent \textbf{Canonical statistical discrimination model}

 \citet{aigner1977statistical} showed that when a rational decision-maker is presented with applicants' noisy unbiased signals for ability,  
 she then has an incentive to weigh applicants' group membership alongside those signals to improve the accuracy of her predictions. As a result, the decision-maker will assign equal outcomes to applicants with equal \textit{expected} ability across groups, but unequal outcomes for applicants with equal \textit{true} ability. The result was foundational in identifying a root cause of discrimination, based not on emotion or taste, but on mathematical judgment.  

Our model studies a generalized form of statistical discrimination.  
We consider cases where the decision-maker accesses an arbitrary set of variables $X$ to predict ability $A$, estimating a signal $S = \bb E [A|X]$   from the econometric model $A = S + \varepsilon$.
We assume that $X$ explains group differences in outcomes. 
This is trivially satisfied if $X$ contains  an unbiased test score and a demographic group identifier, in which case  our model maps  to  \citet{aigner1977statistical}. 
If, however, $X$ contains only non-demographic variables that  explain the variation in  ability across  demographic groups, then the signal $S$ will still exhibit  reversion to group-specific means. 
Therefore, the presence of variables that explain group differences in ability will replicate the same mathematical effect caused by statistical discrimination \citep{aigner1977statistical}.  

As a practical matter, this model shows that while deleting demographic variables from datasets resolves the disparities in \citet{aigner1977statistical}, it 
is not a reliable solution to achieve fair prediction in general.
Instead, disparities can be eliminated by increasing the accuracy of predictions for the affected group: data enrichment, as opposed to data omission.




\section{Empirical application}

\begin{figure}
  \centering
  \includegraphics[width=\textwidth]{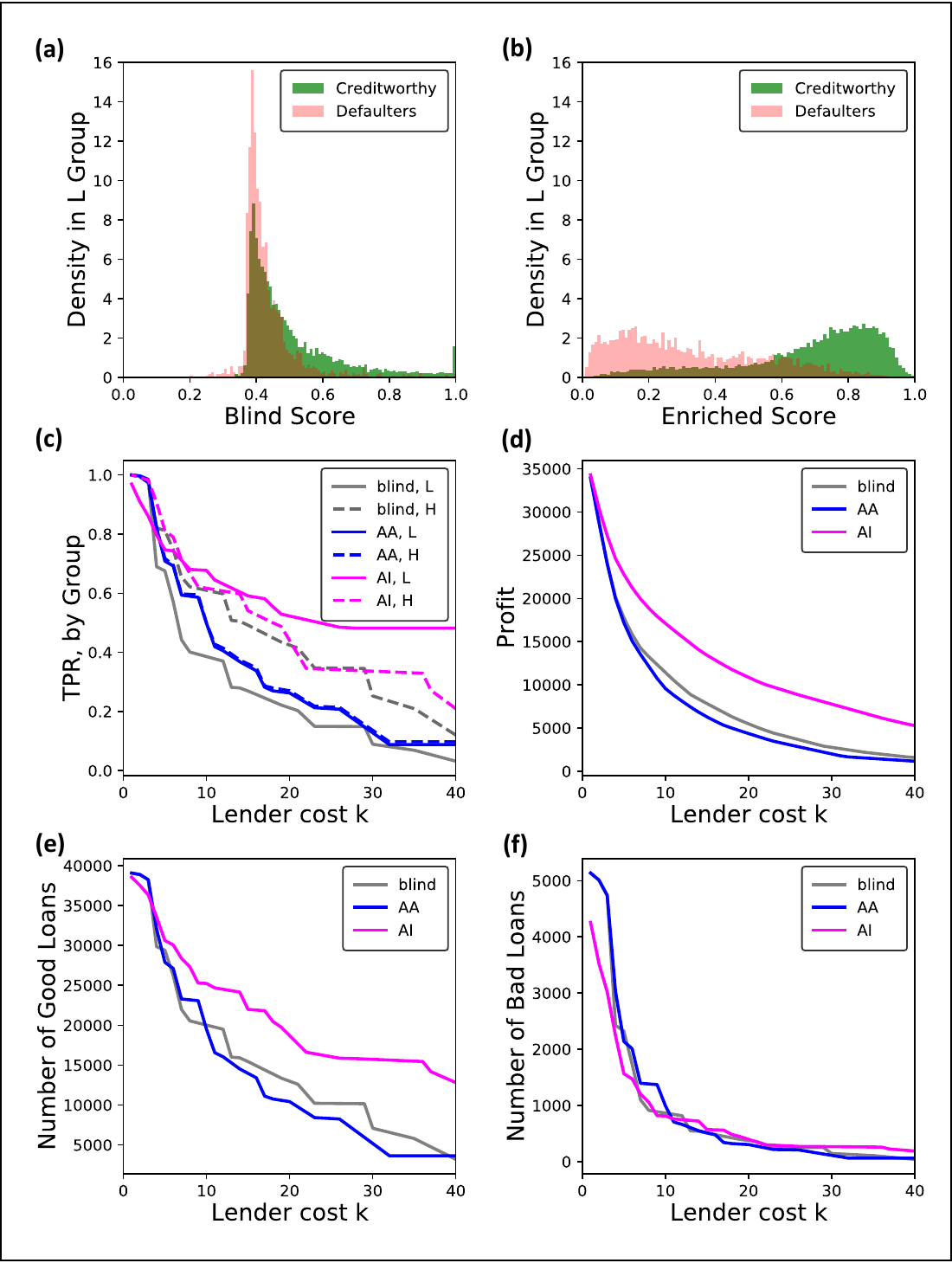} 
  \caption{\label{emp-fig}  \small (a) Scores based on limited dataset are bunched at low values for creditworthy and defaulting applicants. (b) Scores based on enriched dataset, in Affirmative Information scheme, better distinguish creditworthy applicants. (c) When  blind prediction (gray) is replaced by Affirmative Action (blue), the TPR of the higher-mean group falls and the TPR of the lower-mean group increases. Meanwhile, when blind prediction is replaced by Affirmative Information (pink), the TPR of the higher-mean group is roughly unchanged and the TPR of the lower-mean group increases. For all but the lowest values of $k$, Affirmative Information is associated with higher TPRs for both groups than Affirmative Action. (d) Affirmative Information corresponds to greater lender profits than both blind prediction and Affirmative Action. (e) Affirmative Action tends to reduce the total number of good loans administered compared to blind prediction, whereas Affirmative Information increases that number beyond the levels in both alternative schemes. (f) All three schemes yield similar numbers of bad loans administered.}
\end{figure}

We now return to our empirical example from Section \ref{motiv}, where we demonstrated how a set of credit risk scores can lead to disparate outcomes across education groups. Despite the fact that the scores were constructed without group identifiers or sensitive predictors, they yielded lower true and false positive rates to less-educated applicants at every cutoff.  Here, we contrast traditional group-blind prediction with two strategies for achieving equal opportunity for  creditworthy individuals: Affirmative Action and Affirmative Information. Affirmative Action works by adjusting the disadvantaged group's criteria for acceptance, while Affirmative Information works by adding data to enhance the accuracy of the disadvantaged group's predictions. 

As before, we use  SIPP, a nationally-representative
panel survey of the civilian population \citep{sipp_cite}. We label our outcome of ``creditworthiness" as the successful payment of rent, mortgage, and utilities in every month of 2014. 
Of the overall sample, we call the 55\% who studied beyond a high school education group $H$ and those with at most a high school education group $L$. 

We suppose a lender is using repayment predictions to assign loan denials and approvals, without observing individuals' group membership. The lender approves loans for anyone whose score corresponds to a positive expected return, and therefore imposes a cutoff rule that depends on the relative cost of false positive and false negative errors. 
To model the lender choice, we consider a profit function that generates cutoff decision rules: $\text{TP} - k \text{FP}$, where $\text{TP}$ is the number of creditworthy applicants given loans, $\text{FP}$ is the number of defaulters given loans, and $k$ is a constant representing the relative cost of false positive classifications.  We compute the lender's classification choices  by iteratively considering different possible values of $k$ in the profit function, and determining at each $k$ what would be the lender's endogenously chosen classification decisions.

The lender first formulates group-blind predictions of repayment ability using a logistic regression on total assets, debt, and income. In the blind scheme, they select a single profit-maximizing cutoff without regard to group membership. In an Affirmative Action scheme, meanwhile, they  choose an accuracy-maximizing pair of group-dependent cutoffs among the set that generates equal TPRs among $L$ and $H$ applicants. Lastly, in the Affirmative Information scheme, we suppose the decision-maker has taken effort to gather additional data about the $L$ applicants, and to simulate that we train new scores for $L$ applicants using the much richer set of variables available in SIPP, spanning more detailed household financials as well as  sensitive characteristics such as food security, family structure, and participation in social programs. The lender then makes profit-maximizing acceptance decisions based off the original  $H$ scores and the enriched $L$ scores.
 To summarize the difference between the schemes, Affirmative Action works by enforcing more lenient cutoffs for applicants in $L$, while Affirmative Information works by enhancing the accuracy of $L$ scores.
The distribution of $L$ scores used in the Affirmative Action scheme are depicted in Figure \ref{emp-fig}(a) while those used in the Affirmative Information scheme are depicted in (b).

In Figure \ref{emp-fig}(c), we plot creditworthy applicants' access to positive classifications given different values of lender cost $k$. The gray lines show that blind prediction yields consistently higher TPRs for members of $H$ compared to $L$, as in Section \ref{motiv}. The Affirmative Action intervention is depicted in blue and the Affirmative Information intervention in pink. We see that Affirmative Action eliminates the disparity generated by blind prediction, but does not push the TPR for $L$ all the way to the original levels of the $H$ applicants. Rather, the added cost to the lender of reducing disparities leads them to set cutoffs that cause the two groups' TPRs to meet in the middle,  raising the TPR of $L$ while also reducing that of $H$. By comparison, Affirmative Information is associated with leaving the original TPR for $H$ essentially unchanged, while significantly raising the TPR for $L$ beyond its level in blind prediction or Affirmative Action for all but the smallest values of $k$. Compared to Affirmative Action, it yields greater access to positive classifications for both groups' creditworthy applicants. 

Figure \ref{emp-fig}(d) illustrates the impact on lender profit from each intervention. For many values of $k$, Affirmative Action reduces profit. Yet for all values of $k$, Affirmative Information increases profit. In Figure \ref{emp-fig}(e) and (f), we see the effect works through increasing the overall number of repaid loans, while leaving the overall number of defaults essentially unchanged.

There are two noteworthy caveats to Affirmative Information. The first is that Affirmative Information requires data acquisition, which is  costly and will reduce profits below the level seen in Figure \ref{emp-fig}(d). The second is that it may be less effective in lemon-dropping settings in which cutoffs are well below the mean, as seen in Figure \ref{emp-fig}(c) at the lowest values of $k$. 

Overall, this empirical example has demonstrated the potential advantages of Affirmative Information as a tool to broaden access to opportunity.  Particularly in  settings with selective admissions criteria, employing Affirmative Information rather than Affirmative Action can yield TPRs that are higher for \textit{both} groups. In addition, depending on the cost of the data acquisition, Affirmative Information may benefit decision-makers as well. Targeted data enrichment can therefore create natural incentives to broaden access to opportunities.

\section{Conclusion}

This paper has shown that even facially neutral prediction models can create systematic disparities in access to opportunity. 
Even if the models omit variables tracking demographic characteristics and their proxies, uncertainty causes them to assign different probabilities of acceptance to equally-qualified applicants in groups with differing mean ability. 
This effect, which we have called the disparate impact of uncertainty, can arise in all settings where decision-makers screen applicants who come from demographic groups with different risk distributions: 
it can thus compromise fair judgment in  lending, bail decisions, college admissions, hiring, medical treatment assignments, and foster care placements, to name a few examples. 

We hope that our work   will motivate decision-makers to consider shifting their antidiscrimination efforts from traditional data omission to thoughtful data enrichment.  
  While traditional analyses of fair prediction  implicitly treat the available data as  fixed, we explored how relaxing that constraint and   permitting data acquisition  expands the frontier of possible prediction to  achieve outcomes that are both  more efficient and more equitable. 
  
  Now that the Supreme Court has struck down Affirmative Action, Affirmative Information may serve as an effective alternative to expand access to opportunity while supporting competing notions of fairness.
  At least two  roads lie ahead.
  First is statistical guidance for private actors on how they can pinpoint which data would best identify qualified individuals if they were to gather it. 
  Second is how the public, through  economic incentives and the law, can enable private actors to invest in data acquisition. 
  

  \section{Acknowledgments}

This work benefitted from Anna Mikusheva and David Autor's  invaluable advice at each stage. Thank you to them, and to Frank Schilbach, Thomas Brennan, Iván Werning, Victor Chernozhukov, Ben Bernanke, Daron Acemoglu,    Suhas Vijaykumar, Ben Deaner, David Hughes,   Jacob Goldin,   Mitch Polinsky, Sarath Sanga, 
 Daniel Ho, Mark Kelman,  Rahul Singh, David Sontag,  Joseph Doyle,  and Mert Demirer. 
 
 This work was supported by the Arthur and Toni Rembe Rock Center for Corporate Governance at Stanford,   the John M. Olin Program in Law and Economics at Stanford, and the National Science Foundation Graduate Research Fellowship under Grant No. 1122374.
  
  \section{Disclosure statement}
  
  The author reports there are no competing interests to declare.

\newpage
\bibliographystyle{apalike}
\bibliography{draft}

\section*{Appendix for Section 3.1}

 \doublespacing

\noindent
\textbf{Preliminary result supporting proof of Lemma 1}

\begin{lemma} \label{overrep-lemma}
$L$ members are \textit{not} over-represented among positive classifications if and only if 
\begin{equation} \label{app-lemma-1}
\frac{\hat \mu_H }{\hat \mu_L} \geq  \frac{\mu_H  }{\mu_L},
\end{equation}
where $\hat \mu_G = \mathbb{P}[\hat y =1 | G] $ are the group-specific positive classification rates.
Furthermore, if $L$ members are not over-represented, then it follows that 
\begin{equation} \label{app-lemma-2}
\hat \mu_H > \hat \mu_L
\end{equation} 
\end{lemma}
\begin{proof}
We prove $\mathbb{P}[G=L | \hat y =1] \leq \mathbb{P}[G=L | Y =1] \iff (\ref{app-lemma-1})$ by observing the equivalence of the following statements. We will use Bayes' rule, as well as the fact that $\hat \mu_G \in (0,1)$ given the assumption that each groups' true and false positive rates $\text{TPR}_G, \text{FPR}_G \in (0,1)$ (see footnote 6 in the main text).
\begin{align*}
\mathbb{P}[G=L | \hat y =1] &\leq \mathbb{P}[G=L | Y =1]  \\[5pt] 
\frac{\mathbb{P}[\hat y =1 | G=L] \mathbb{P}[G=L]}{\mathbb{P}[\hat y =1]} &\leq \frac{\mathbb{P}[ Y =1 | G=L] \mathbb{P}[G=L]}{\mathbb{P}[Y =1]} \\[5pt] 
\frac{\mathbb{P}[\hat y =1 | G=L]}{\mathbb{P}[\hat y =1]} &\leq \frac{\mathbb{P}[ Y =1 | G=L]}{\mathbb{P}[Y =1]} \\[5pt] 
\frac{\hat \mu_L}{\hat \mu_L \mathbb{P}[G=L]+\hat \mu_H \mathbb{P}[G=H]} &\leq \frac{\mu_L}{\mu_L \mathbb{P}[G=L] +\mu_H \mathbb{P}[G=H] } \\[7pt] 
\frac{\hat \mu_L \mathbb{P}[G=L]+\hat \mu_H \mathbb{P}[G=H]}{\hat \mu_L} &\geq \frac{\mu_L \mathbb{P}[G=L] +\mu_H \mathbb{P}[G=H] }{\mu_L} \\[5pt] 
\mathbb{P}[G=L] + \frac{\hat \mu_H \mathbb{P}[G=H]}{\hat \mu_L} &\geq \mathbb{P}[G=L] + \frac{\mu_H \mathbb{P}[G=H] }{\mu_L} \\[5pt]
\frac{\hat \mu_H \mathbb{P}[G=H]}{\hat \mu_L} &\geq  \frac{\mu_H \mathbb{P}[G=H] }{\mu_L} \\[5pt] 
\frac{\hat \mu_H }{\hat \mu_L} &\geq  \frac{\mu_H  }{\mu_L} 
\end{align*}
And from this and the assumption that $\mu_H>\mu_L$, the inequality (\ref{app-lemma-2}) also follows. 
\end{proof}

\noindent
\textbf{Detailed proof of Lemma 1}

\noindent
If (i) holds, then from Lemma \ref{overrep-lemma}, we know
\begin{equation} \label{overrep-cond-rewritten-appendix}
 \frac{\mu_H  }{\mu_L} \leq \frac{\hat \mu_H }{\hat \mu_L},
\end{equation}
and $\hat \mu_L, \hat \mu_H \in (0,1)$. 

\noindent
Now suppose (ii) also holds, separately considering two cases: $\text{PPV}_L < \text{PPV}_H$ and $\text{NPV}_L \leq \text{NPV}_H$.

Case 1: Rewriting (\ref{overrep-cond-rewritten-appendix}) gives $\frac{\hat \mu_L}{\mu_L}  \leq   \frac{\hat \mu_H}{\mu_H}$, and combining with $\text{PPV}_L < \text{PPV}_H$, we get 
\begin{equation*} \label{tpr-bayes-appendix}
 \text{PPV}_L \frac{\hat \mu_L}{\mu_L} <  \text{PPV}_H \frac{\hat \mu_H}{\mu_H}. 
\end{equation*} 
Note that this is  $\text{TPR}_L < \text{TPR}_H$ rewritten with Bayes' rule, since 
\begin{equation*}
\text{TPR}_G = \bb P [\hat y =1 | Y=1 , G] = \frac{\bb P [Y=1 | \hat y=1, G] \bb P [\hat y =1 |G]}{\bb P [Y=1|G]} \equiv \text{PPV}_G  \frac{\hat \mu_G}{\mu_G}. 
\end{equation*}

Case 2: Note that  (\ref{overrep-cond-rewritten-appendix}) implies $\hat \mu_L < \hat \mu_H$, since $\mu_L < \mu_H$. Combining with $\text{NPV}_L \leq \text{NPV}_H$,  
\begin{equation}
(1-\text{NPV}_L) \frac{1-\hat \mu_L}{\mu_L} > (1-\text{NPV}_H) \frac{1-\hat \mu_H}{\mu_H}. \label{fnr-bayes-appendix} 
\end{equation}
We can see this is $\text{FNR}_L > \text{FNR}_H$ rewritten with Bayes' rule, where $\text{FNR}_G$ is the group-specific false negative rate $\bb P [\hat y = 0 | Y=1, G]$. In particular, 
\begin{align*}
\text{FNR}_G  = \bb P [\hat y = 0 | Y=1, G] &= \frac{\bb P[Y=1 | \hat y =0 , G] \bb P[\hat y =0 |G ]}{\bb P [Y=1|G]} \\ &=  \frac{(1- \bb P[Y=0 | \hat y =0 , G])(1-\bb P[\hat y =1 |G ])}{\bb P [Y=1|G]} \\
&\equiv (1- \text{NPV}_G) \frac{1 - \hat \mu_G}{\mu_G}.
\end{align*}

The groups' false negative  and true positive rates are related by $\text{FNR}_G = 1 - \text{TPR}_G$ because 
$\text{FNR}_G = \bb P [\hat y = 0 | Y=1, G] = 1- \bb P [\hat y = 1 | Y=1, G] = 1- \text{TPR}_G$.
Therefore the inequality (\ref{fnr-bayes-appendix}) gives $1- \text{TPR}_L > 1- \text{TPR}_H \implies \text{TPR}_L < \text{TPR}_H$.

\vspace{1cm}
\noindent
\textbf{Characterizing how to achieve equality in all group error rates (not just TPRs)}

In order to achieve balance in both error rates, the TPR and FPR, members of $L$ must be over-represented among the positive classifications. In particular, $L$ must comprise a strictly greater portion of the positive classifications $\hat y =1$ than it does positive types $Y=1$. This is formally stated and proven below.

 \begin{proposition} \label{over-representation}
If $\hat y$ yields $\text{TPR}_L \geq \text{TPR}_H $ and $\text{FPR}_L \geq \text{FPR}_H $
then it must be that $L$ members are over-represented among positive classifications. That is, $\mathbb{P}[G=L | \hat y =1] > \mathbb{P}[G=L | Y =1]$.
\end{proposition}  

\begin{proof}
Proof by contradiction. Suppose that $L$ members are \textit{not} over-represented among positive classifications.

Thus, by Lemma \ref{overrep-lemma}, 
\begin{equation} \label{class-rate-ineq-2}
\frac{\hat \mu_H }{\hat \mu_L} \geq  \frac{\mu_H  }{\mu_L}
\end{equation}
and  
\begin{equation} \label{class-rate-ineq-1}
\hat \mu_H > \hat \mu_L. 
\end{equation} 

We will use (\ref{class-rate-ineq-2}) and (\ref{class-rate-ineq-1}) to arrive at a contradiction when the TPRs and FPRs  are at least as great for the $L$ group as for the $H$ group. First, using Bayes' rule, we rewrite the following group-specific error rates in terms of the group-specific positive predictive values $PPV_A $, rates of positive classification $\hat \mu_A$, and base rates $\mu_A$. 
\begin{align}
&\text{TPR}_G = \mathbb{P}[\hat y =1 | G, Y=1] 
= \frac{\mathbb{P}[Y=1|G, \hat y =1] \mathbb{P}[\hat y =1 |G]}{\mathbb{P}[Y=1 | G]} 
= \text{PPV}_G \frac{\hat \mu_G}{\mu_G} \nonumber \\
&\text{FPR}_G = \mathbb{P}[\hat y =1 | G, Y=0] = \frac{\mathbb{P}[Y=0|G, \hat y =1] \mathbb{P}[\hat y =1 |G]}{\mathbb{P}[Y=0 | G]} 
= (1-\text{PPV}_G) \frac{\hat \mu_G}{1-\mu_G} \nonumber
\end{align}
Then for the $\text{FPR}_L$ to be at least as big as $\text{FPR}_H$, it must be that 
\begin{align}
(1-\text{PPV}_L) \frac{\hat \mu_L}{1-\mu_L}  &\geq (1-\text{PPV}_H) \frac{\hat \mu_H}{1-\mu_H} \nonumber \\[5pt]
\frac{1-\text{PPV}_L}{1-\text{PPV}_H} &\geq \frac{\hat \mu_H}{\hat \mu_L} \frac{(1-\mu_L)}{(1-\mu_H)} \label{fpr_inequality}
\end{align}
According to (\ref{class-rate-ineq-1}) and the assumption $\mu_L<\mu_H$, the RHS of (\ref{fpr_inequality}) is strictly greater than 1. Therefore it must be that 
\begin{equation} \label{ppv-less}
\text{PPV}_L < \text{PPV}_H.
\end{equation}
Meanwhile, for $\text{TPR}_L$ to be at least as big as $\text{TPR}_H$, 
\begin{align}
\text{PPV}_L \frac{\hat \mu_L}{\mu_L}  &\geq \text{PPV}_H \frac{\hat \mu_H}{\mu_H} \nonumber \\[5pt]
\frac{\text{PPV}_L}{\text{PPV}_H} &\geq \frac{\hat \mu_H}{ \mu_H} \frac{\mu_L}{\hat \mu_L} \nonumber \\[5pt]
\frac{\text{PPV}_L}{\text{PPV}_H} &\geq \frac{\frac{\hat \mu_H}{\hat \mu_L}}{\frac{\mu_H}{\mu_L}} \nonumber
\end{align}
Due to (\ref{class-rate-ineq-2}) and the assumption $\mu_L < \mu_H$, the RHS is weakly greater than 1. Therefore, it must be that $\text{PPV}_L \geq \text{PPV}_H$ which contradicts with (\ref{ppv-less}). 
\end{proof}

\section*{Appendix for Section 3.2}

\noindent
\textbf{Proving that scores are best prediction regardless of whether group membership observed}

We can prove that those scores continue to serve the decision-maker as the best prediction regardless of whether group membership is observed: 
\begin{proposition} \label{mpc}
The lender's risk scores $S = \bb E [A|X]$ will satisfy $\bb E [A |S, G] = S$. That is, an applicants' expected ability given their score will equal their score, regardless of group membership.
\end{proposition}
We know from Proposition \ref{mpc} that the lender's best available prediction of an applicant's true ability is given to her by the score. If she imposes a cutoff rule to classify individuals, then she will choose the same cutoff for each group regardless of whether she observes group affiliation. 

\begin{proof}
We will use a lemma in Mitchell et al. (2019), based on the law of iterated expectations: any three random variables $W, V, Z$ satisfy $\bb E [W | V, Z] = \bb E [W | \bb E [W|V,Z],Z] $. 

Applied to our setting, we have
\begin{align}
\bb E[A|X, G] &= \bb E[A| \bb E [A| X,G], G]  \text{ by lemma} \nonumber \\
  &= \bb E[A| \bb E [A| X], G]  \text{ by mean independence } \bb E [A|X,G] = \bb E[A|X] \nonumber \\
  &= \bb E[A| S, G] \text{ by definition of } S \nonumber
\end{align}

From our conditional mean independence assumption, we also know that the LHS is equal to $\bb E [A|X] = S$. Combining with the RHS, we have $S = \bb E[A|S,G]$.

\end{proof}

\noindent
\textbf{Derivation of $\bb E[S_g|A_g]$}

To compute the expected score, we use the fact that when variables $X_1$ and $X_2$ are bivariate normally distributed, then the distribution of $X_1$ given $X_2=x_2$ has conditional expectation $\bb E[X_1] + \frac{\text{cov}(X_1,X_2)}{\sigma_{X_2}^2}   (x_2 - \bb E[X_2])$. In our model, then, for both $G \in \{ L, H\},$ 
\begin{align*}
\bb E[S_G | A_G = a]  &= \bb E[S_G] + \frac{\text{cov}(S_G,A_G)}{\sigma_{A_G}^2}  (a - \bb E[A_G]) \\[5pt]
& = \mu_G + \frac{\text{cov}(S_G,S_G+\varepsilon_G)}{\sigma_{S_G}^2+ \sigma_{\varepsilon_G}^2}   (a - \mu_G) \\[5pt]
&= (1-\gamma_G) \mu_G + \gamma_G a \text{, where } \gamma_G \equiv    \frac{\sigma_{S_G}^2}{\sigma_{S_G}^2+ \sigma_{\varepsilon_G}^2} 
\end{align*}

\noindent
\textbf{Derivation of $S|A>0$}

We can derive the mean and variance for the distribution of $X | Y> c $ where $(X, Y)$ are bivariate normal. To find the mean, use the law of iterated expectations:
\begin{align}
\bb E [X | Y> c] = \bb E [\bb E [X | Y ] |Y>c]  \nonumber
\end{align} 

The inner expectation $E [X | Y = y] $ is: 
\begin{align}
\bb E [X | Y = y ] = \mu_x + \text{cov}(X,Y) (\frac{y- \mu_y}{\sigma^2_y})  \nonumber
\end{align} 

Take the expectation of this given $Y>c$:
\begin{align}
\bb E [\bb E [X | Y ] |Y>c]  &= \mu_x + \text{cov}(X,Y) (\frac{\bb E [Y | Y>c] - \mu_y}{\sigma^2_y}) \nonumber \\
&= \mu_x + \text{cov}(X,Y) \frac{(\mu_y + \sigma_y (\frac{\phi(\frac{c-\mu_y}{\sigma_y})}{1-\Phi(\frac{c-\mu_y}{\sigma_y})})) - \mu_y}{\sigma^2_y} \nonumber 
\end{align} 

which simplifies to
\begin{align}
  &\mu_x + \text{cov}(X,Y) \frac{\phi(\frac{c-\mu_y}{\sigma_y})}{\sigma_y(1-\Phi(\frac{c-\mu_y}{\sigma_y}))} \nonumber \\
= &\mu_x +  \frac{\rho \sigma_x \phi(\frac{c-\mu_y}{\sigma_y})}{1-\Phi(\frac{c-\mu_y}{\sigma_y})} \nonumber
\end{align} 

And we get
\begin{align}
\bb E [X|Y>c ] &= \mu_x + \frac{\rho \sigma_x \phi(\frac{c-\mu_y}{\sigma_y})}{1-\Phi(\frac{c-\mu_y}{\sigma_y})} \nonumber
\end{align}

The variance can be computed similarly, using $\text{Var}(X|Y>c) = \bb E[X^2|Y>c] - \bb E [X|Y>c]^2$. Solving and simplifying with Mathematica gives:
\begin{align}
\text{Var}[X|Y>c] &= \sigma_x^2 (1-\frac{\rho^2 \phi (\frac{c-\mu_y}{\sigma_y})[\sigma_y \phi(\frac{c-\mu_y}{\sigma_y})-(c-\mu_y)(1-\Phi(\frac{c-\mu_y}{\sigma_y})) ]}{\sigma_y(1-\Phi(\frac{c-\mu_y}{\sigma_y}))^2}) \nonumber
\end{align}

\end{document}